\documentclass[conference]{IEEEtran}

\usepackage{array}
\usepackage{subcaption}
\usepackage{graphicx}
\usepackage{wrapfig}
\usepackage{cite}
\usepackage{float}
\usepackage{hyperref}
\usepackage{color}
\usepackage{url}
\usepackage{enumitem}
\usepackage{booktabs}
\usepackage{multirow}
\usepackage[table,xcdraw]{xcolor}
\usepackage{mathtools}
\usepackage{rotating}

\begin{document}

\title{Multiple Sclerosis Lesion Analysis in Brain Magnetic Resonance Images: Techniques and Clinical Applications}

\author{\IEEEauthorblockN{Yang Ma\IEEEauthorrefmark{1},
Chaoyi Zhang\IEEEauthorrefmark{1},
Mariano Cabezas\IEEEauthorrefmark{2},
Yang Song\IEEEauthorrefmark{3},
Zihao Tang\IEEEauthorrefmark{1}\IEEEauthorrefmark{2}, \\
Dongnan Liu\IEEEauthorrefmark{1}\IEEEauthorrefmark{2},
Weidong Cai\IEEEauthorrefmark{1},
Michael Barnett\IEEEauthorrefmark{2}\IEEEauthorrefmark{4},
Chenyu Wang\IEEEauthorrefmark{2}\IEEEauthorrefmark{4}}

        \IEEEauthorblockA{\IEEEauthorrefmark{1}School of Computer Science, University of Sydney, Australia}
        \IEEEauthorblockA{\IEEEauthorrefmark{2}Brain and Mind Centre, University of Sydney, Australia}
        \IEEEauthorblockA{\IEEEauthorrefmark{3} School of Computer Science and Engineering, University of New South Wales, Australia}
        \IEEEauthorblockA{\IEEEauthorrefmark{4}Sydney Neuroimaging Analysis Centre, Australia}
        \IEEEauthorblockA{Contact e-mail: \href{mailto:yama5878@somewhere.com}{\{yama5878, czha5168, ztan1463\}@uni.sydney.edu.au}, \\ \href{mailto:yang.song1@unsw.edu.au}{yang.song1@unsw.edu.au}, 
        \href{mailto:mariano.cabezas@sydney.edu.au}{\{mariano.cabezas, dongnan.liu, tom.cai\}@sydney.edu.au}, \\
        \href{mailto:tim@snac.com.au}{tim@snac.com.au}, \href{mailto:michael@sydneyneurology.com.au}{michael@sydneyneurology.com.au}}
}

\maketitle

\begin{abstract}
Multiple sclerosis (MS) is a chronic inflammatory and degenerative disease of the central nervous system, characterized by the appearance of focal lesions in the white and gray matter that topographically correlate with an individual patient’s neurological symptoms and signs. Magnetic resonance imaging (MRI) provides detailed in-vivo structural information, permitting the quantification and categorization of MS lesions that critically inform disease management. Traditionally, MS lesions have been manually annotated on 2D MRI slices, a process that is inefficient and prone to inter-/intra-observer errors. Recently, automated statistical imaging analysis techniques have been proposed to detect and segment MS lesions based on MRI voxel intensity. However, their effectiveness is limited by the heterogeneity of both  MRI  data acquisition techniques and the appearance of MS lesions. By learning complex lesion representations directly from images, deep learning techniques have achieved remarkable breakthroughs in the MS lesion segmentation task. Here, we provide a  comprehensive review of state-of-the-art automatic statistical and deep-learning MS segmentation methods and discuss current and future clinical applications. Further, we review technical strategies, such as domain adaptation, to enhance MS lesion segmentation in real-world clinical settings.

\end{abstract}



\section{Introduction}

Multiple sclerosis (MS) is a chronic inflammatory and degenerative disease of the central nervous system (CNS) with protean neurological manifestations that reflect both focal and diffuse damage to the brain and spinal cord. Visual, motor, sensory, and sphincter dysfunction are common clinical features, which in early disease usually follow a relapsing course with partial remissions; untreated, the majority of patients will develop progressive motor and cognitive disability~\cite{coles2008alemtuzumab}. The disease is relatively common in Caucasian populations, and both genetic and environmental factors, culminating in immune dysregulation and CNS inflammation, have been causally implicated~\cite{jh2000lucchinetti}. The prevalence and incidence of MS are increasing globally, particularly in females, for unknown reasons~\cite{world2008multiple}. Globally there are an estimated $2.3$ million people with MS, and after trauma, the disease constitutes the most common cause of neurological disability in young adults.

Magnetic resonance imaging (MRI) sensitively detects MS lesions and is a crucial tool for the diagnosis of MS, objective monitoring of disease progression, and the assessment of treatment efficacy~\cite{polman2011diagnostic}. In clinical trials of putative therapeutics, MRI-derived lesion metrics (number, volume, type) are critical endpoints~\cite{pontillo2021radiomics}.  As such, the segmentation and quantification of MS lesions require high precision and accuracy.  More recently, the advent of precision medicine has seen these metrics emerge as novel tools for informing the management of individual patients in routine clinical practice~\cite{chitnis2020roadmap}.

MS lesions are commonly annotated and measured manually by radiologists or trained neuroimaging analysts with semi-automated annotation tools in clinical trials and imaging studies. Although manual segmentation is still essential, it is impractical for broader clinical applications, since it is generally time-consuming and labor-intensive, and suffers from inter-/intra-observer biases and errors~\cite{commowick2018objective}. As such, there is a growing need for fully automated tools that provide accurate and precise segmentation of MS lesions rapidly from MRI~\cite{llado2012automated}. A number of automatic segmentation methods have been proposed, and the recent incorporation of deep learning methods has rapidly accelerated the development of more accurate automatic lesion segmentation tools~\cite{ravi2016deep,zhang2020multiple}.  However, challenges remain and current state-of-the-art techniques remain inferior to expert manual segmentation~\cite{wattjes20212021}.

One of the more common problems with existing machine learning methods is poor generalization in the context of multi-center studies. Models that are trained with images from one scanner may yield poor results when tested with images from another scanner or acquired with a different protocol~\cite{guan2021dasurvey}. \textcolor{black}{Furthermore, the Dice similarity coefficient (DSC), a common evaluation metric for lesion segmentation, can provide a biased estimation of the segmentation accuracy due to the influence of large lesions. Consequently, acceptable average values for a set of subjects, might shadow a poor performance on a subset of cases with small lesions~\cite{carass2020evaluating}.} Additionally, the spatial distribution and appearance of MS lesions is heterogeneous~\cite{polman2011diagnostic}. Finally, MS lesions constitute only a small fraction of the total brain volume, resulting in a highly imbalanced dataset that poses further challenges to train models~\cite{valverde2017improving}. In this survey, we provide a comprehensive review of the most recently described MS segmentation techniques, to assist imaging researchers with the development of robust, precise, and accurate lesion segmentation solutions for real-world clinical applications.


There are several previous surveys of MS lesion segmentation techniques, the majority of which were published before 2013 and focused on statistical-modeling segmentation methods~\cite{mortazavi2012segmentation,llado2012segmentation,llado2012automated,garcia2013review}. Danelakis et al.~\cite{danelakis2018survey} reviewed MS segmentation methods published between 2013 to 2018; and Kaur et al.~\cite{kaur2020state} included three methods described in 2019 in addition to earlier methods. Finally, Zhang et al.~\cite{zhang2020multiple} reviewed deep-learning methods up to 2020. However, a combined analysis of both statistical and deep-learning methods is lacking, as is a comprehensive discussion of their clinical integration and application.

In this work, we present a comprehensive survey of MS lesion segmentation methods published prior to 2021, unifying previous statistical modeling approaches with state-of-the-art deep-learning frameworks. We specifically discuss the potential for these segmentation methods to address an urgent clinical need for quantitative lesion activity analysis (the measurement of longitudinal lesion change); and also review domain adaptation solutions to lesion segmentation in multicenter, multi-scanner user
 scenarios~\cite{mortazavi2012segmentation,llado2012segmentation,llado2012automated,garcia2013review,danelakis2018survey,kaur2020state,zhang2020multiple}.

{\color{black}{
The main review body is organized as follows. First, we introduce the general biomedical background of MS lesion imaging in Section~\ref{sec-bg-les}, followed by an in-depth survey on the recent MS lesion segmentation methods based on statistical analysis and deep learning (Section~\ref{sec-les-seg}). We further discuss lesion activity analysis techniques in Section~\ref{sec-les-app}, together with domain adaptation strategies to improve model robustness across different scanners. In Section~\ref{sec-evaluation}, we summarize popular public datasets, evaluation metrics, and state-of-the-art methods for MS lesion segmentation. Finally, we conclude the overall paper in Section~\ref{sec-conclusion} and discuss promising research opportunities together with their challenges.
}}



\section{Imaging the MS lesion  \label{sec-bg-les}}

\begin{figure}
 \centering
 \includegraphics[scale=0.85]{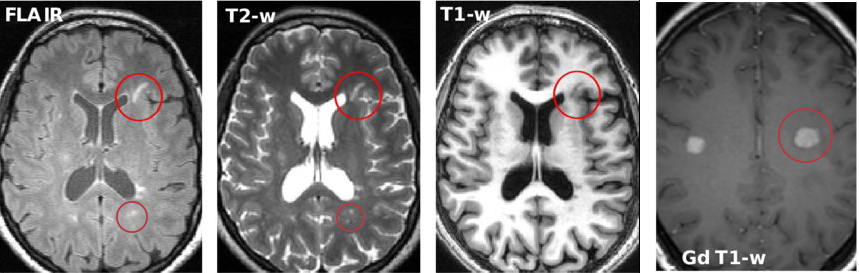}
 \caption{T2-FLAIR, T1-w, T2-w and Gd T1-w images of brain damage. MS lesions are indicated by red circles.}
 \label{fig:intro1}
\end{figure}

To visualize MS lesions, four imaging sequences are commonly acquired: PD-weighted (PD-w), T1-weighted (T1-w), T2-weighted (T2-w), and T2 fluid-attenuated inversion recovery (T2-FLAIR) imaging~\cite{hashemi2012mri}, as shown in Fig.~\ref{fig:intro1}. Generally, structural T1-w and FLAIR are most frequently used in clinics and contain all necessary information to identify MS lesions with modern 1.5 and 3.0T scanners, while PD-w and T2-w images offer superior visualization of lesions in the posterior fossa.

Compared with  {\textcolor{black}{normal appearing white matter}}, MS lesions appear hypointense on T1-w images; and are usually hyperintense on T2-w, PD-w, and FLAIR images (Fig.~\ref{fig:intro1}).  The identification of lesions on PD-/T2-w imaging is hampered by the similarity of cerebrospinal fluid (CSF) and lesional voxel intensities, particularly in the commonly involved peri-ventricular white matter zone. T2-FLAIR imaging, which attenuates CSF signal, permits hyperintense MS lesions to be seen to a better advantage. Therefore, a multimodal approach provides the best opportunity to robustly and accurately segment MS lesions.

MS lesions are heterogeneous both in morphology and signal characteristics.  Most commonly, MS lesions appear as small ovoid hyperintensities on T2-w, PD-w, and FLAIR images; and are hypointense on T1-w images.  Lesion appearance may evolve serial imaging, due to changes in lesion water content, myelin and axon density, inflammatory cell populations, and gliosis.  Rarely, highly inflammatory lesions may liquify, developing a central necrotic zone with similar voxel intensities to CSF. As a consequence, they may appear hypointense on FLAIR-imaging and be missed by models primarily trained with hyperintense lesion masks. Necrotic lesions in the periventricular lesions may be absorbed into the CSF, rendering them undetectable by both conventional and model-derived image analyses.

According to the current revision of diagnostic criteria for MS patients~\cite{polman2011diagnostic}, dissemination of lesion in time (DIT) and space (DIS) are essential for the diagnosis of MS.



DIT can be inferred from a single scan by the simultaneous presence of T2-w/PD-w hyperintense lesions that exhibit gadolinium contrast-enhancement on T1-w imaging, indicative of recent inflammatory activity and associated breakdown of the blood-brain barrier, and non-enhancing lesions. DIT can also be determined by the presence of new MS lesions on a subsequent (follow-up) MRI scan.

DIS is assessed by detecting at least two of the four MS-typical lesion subtypes (juxtacortical, cortical, periventricular or infratentorial, or spinal cord) in a single scan:

\begin{itemize}
    \item \textit{Juxtacortical lesions} are usually small, ellipsoid, or linear in appearance, and located adjacent to the cerebral cortex. However, their small size hinders detection, especially when images are acquired with a slice thickness greater than the lesion size (due to partial volume). High resolution gapless 3D FLAIR images, routinely acquired in many larger MS centers, partially alleviate this issue.
    \item \textit{Cortical lesions} occur within the cortex; while commonly identified in histopathological MS tissue, they are rarely detected in-vivo with conventional clinical scanning protocols. Non-conventional techniques such as double inversion recovery imaging, which suppresses both CSF and  {\textcolor{black}{normal appearing white matter}} signal, may improve cortical lesion detection and segmentation~\cite{gabr2016automated}. Additionally, MP2RAGE, a novel T1-w sequence, is reported to be highly sensitive for cortical lesions~\cite{la2020multiple}.
    \item \textit{Periventricular lesions} are usually 5-10 mm corona-like structures that involve the white matter adjacent to the ventricular CSF. While relatively easy to detect with FLAIR imaging, their heterogeneous shape and intensity make the delineation of their contours a difficult task.
    \item \textit{Infratentorial lesions}, which involve the brainstem or cerebellum lesions, are usually small and may be difficult to discern on FLAIR imaging. Visualized to better advantage with T2-w imaging, infratentorial lesions are frequently symptomatic and may portend a worse prognosis. Accurate segmentation and quantitation may therefore constitute an important and clinically relevant biomarker. A discussion of spinal cord lesions is outside the scope of this review.

\end{itemize}

In previous reviews, discussion of lesion identification techniques for different lesion subtypes was limited, and primarily focused on the segmentation of white matter hyper-intensities (regardless of their subtype); or cortical lesions. \textcolor{black}{Nevertheless, the newer methods that we reviewed for this survey have continued to focus only on hyperintense and cortical lesions, leaving room for improvement when it comes to identifying lesion subtypes.}


\section{Lesion Segmentation Framework~\label{sec-les-seg}}


Several challenges arise when developing segmentation models for MRI data acquired from clinical sites due to intensity inhomogeneity~\cite{roura2015toolbox,santos2016evaluation} and imaging noise~\cite{beaumont2016automatic,valverde2016multiple} inherent to the scanner acquisition. Pre-processing steps are commonly applied to minimize this impact prior to the implementation of the lesion segmentation algorithm. Common pre-processing steps and their 3D representation are summarised as follows:

\begin{itemize}
    \item \textit{Co-registration}~aligns different MRI sequences to the same 3D space in order to benefit from multi-contrast information for lesion segmentation. \textcolor{black}{While multi-modal and longitudinal methods perform rigid transformations followed by image resampling with interpolation~\cite{bardinet2002co,kiebel1997mri}, this pre-processing step is not necessary when dealing with a single image.}
    \item \textcolor{black}{\textit{Denoising} reduces the noise introduced by the acquisition phase, improving the signal-to-noise-ratio, which is commonly used for classical statistical approaches. A wide range of methods, adopted from general computer vision, are used and include the non-local means algorithm~\cite{beaumont2016automatic}, anisotropic diffusion filter~\cite{roura2015toolbox}, and (3D) Gaussian low-pass filter~\cite{knight2016ms,santos2016evaluation}. However, deep-learning approaches are capable of defining useful features for segmentation, and can usually avoid the need of denoising beforehand.}
    \item \textit{Bias correction} ameliorates the intensity inhomogeneities of a given image. Due to imperfections in the acquisition magnet, voxels of the same tissue can have different intensities in different brain regions. Inhomogeneity correction is most commonly performed with the N3 normalisation technique~\cite{roura2015toolbox,santos2016evaluation} and its improvement, the N4 normalisation technique~\cite{beaumont2016automatic,doyle2016automatic,jog2015multi,roy2018multiple,valverde2016multiple}.
    \item \textit{Skull stripping} is used to remove the skull along with other non-brain tissues that may influence the segmentation performance. Two common strategies are template-based and edge-based approaches. Existing toolboxes include VolBrain~\cite{beaumont2016automatic,valverde2016multiple}, {\color{black}FSL's BET~\cite{jenkinson2012fsl,smith2004advances,santos2016evaluation}}, and ROBEX~\cite{iglesias2011robust}.
\end{itemize}



Following image pre-processing steps, both supervised and unsupervised machine learning techniques have been applied to the task of lesion segmentation. The principal difference between these broad categories relates to whether labeled data is used for training. Unsupervised techniques generally aim to simulate the intuitive processes that permit humans with expert domain knowledge, in this case, clinicians and neuroscientists, to accurately segment lesions; and do not rely on annotated data for model training. Using training data labeled by human experts, supervised methods learn the representation of MS lesions using a training process. Training these techniques normally involves explicit feature engineering or the use of deep neural networks. Furthermore, the format of the input data includes 2D-image input (pixel-level) and 3D-volume input (voxel-level).

Supervised methodologies are differentiated into statistical and deep-learning methods:

\begin{itemize}
  \item \textit{Statistical methods} usually apply statistical models to perform probabilistic segmentation on the original data or the hand-crafted features. The engineered features normally incorporate useful interpretation for the MS lesions from the initial input data.
  \item \textit{Deep-learning based methods} use deep convolution neural networks (CNNs) as voxel classification models for segmenting MS lesions. {\color{black}Based on convolutional operations widely used for computer vision tasks on natural images, these methods can extract more salient features and uncover nonlinear intensity relationships between voxels to achieve a greater performance than classical statistical models.}
\end{itemize}

\begin{table*}[t]
  \centering
  \caption{Review of MS segmentation methods. The acronyms for the algorithms stand for (in alphabetical order): Gaussian Mixture Modelling (GMM), Model of Population and Subject (MOPS), and The Rotation-invariant multi-contrast non-local mean segmentation (RMNMS).}
  \label{tab:review}
\begin{tabular}{llllll}
\hline
Method                                                                                         & Supervision  & Strategy      & \begin{tabular}[c]{@{}l@{}}Data \\ Handling\end{tabular} & MRI Data                                                                       & Classifier                                                                        \\ \hline
Weiss et al., 2013~\cite{weiss2013multiple}                              & Unsupervised & Statistical  & 2D input                                                 & T1-w, T2-w                                                                     & \textcolor{black}{Sparse coding}                                                                  \\
Hill et al, 2015~\cite{hill2015automated}                                & Unsupervised & Statistical  & 2D input                                                 & T1-w, T2-w, PD-w                                                               & \textcolor{black}{Jump method clustering}                                                                               \\
Roy et al, 2017~\cite{roy2017effective}                                  & Unsupervised & Statistical  & 2D input                                                 & T1-w, T2-w, PD-w                                                               & Thresholding \\
Tomas-Fernandez et al, 2015~\cite{tomas2015model}                        & Unsupervised & Statistical  & 3D input                                                & T1-w, T2-w, T2-FLAIR                                                           & MOPS                                                            \\

{\color{black}{Roura et al., 2015~\cite{roura2015toolbox}}} & Unsupervised & Statistical  & 3D input      & T1-w, T2-FLAIR & Thresholding \\

{\color{black}{Cerri et al., 2021~\cite{cerri2021contrast}}} & Unsupervised & Statistical & 3D input & T1-w, T1-w Gd, T2-FLAIR & GMM \\

Steenwijk et al., 2013~\cite{steenwijk2013accurate}                      & Supervised   & Statistical & 3D input                                                & T1-w                                                                           & k-NN                                                                              \\
Geremia et al., 2013~\cite{geremia2013spatially}                         & Supervised   & Statistical & 3D input                                                & T1-w, T2-w, T2-FLAIR                                                           & Random Forest  \\

{\color{black}{Guizard et al., 2015~\cite{guizard2015rotation}}} & Supervised & Statistical & 3D input & T1-w, T2-w, T2-FLAIR & RMNMS \\

{\color{black}{Jesson et al., 2015~\cite{jesson2015hierarchical}}} & Supervised & Statistical & 3D input & T1-w, T2-w, T2-FLAIR & Random Forest \\

{\color{black}{Valcarcel et al., 2018~\cite{valcarcel2018mimosa}}} & Supervised & Statistical & 3D input & T1-w, T2-FLAIR, PD-w & Logistic Regression \\

Havaei et al, 2016~\cite{havaei2016hemis}                                & Supervised   & Deep-learning   & 2D input                                                 & T1-w, T2-w, T2-FLAIR                                                           & CNN                                                                               \\

Andermatt et al, 2017~\cite{andermatt2017automated}                      & Supervised   & Deep-learning & 2D input                                                 & T1-w, T2-w, T2-FLAIR, PD-w                                                     & CNN                                                                               \\ 

{\textcolor{black}{Aslani et al., 2019~\cite{aslani2019multi}}}                                & Supervised   & Deep-learning   & 2D input                                                & T1-w, T2-w, T2-FLAIR                                                        & CNN  \\

Ackaouy et al., 2020~\cite{ackaouy2020unsupervised}                      & Supervised   & Deep-learning   & 2D input                                                 & T1-w, T2-w, T2-FLAIR                                                           & CNN  \\

{\color{black}{Birenbaum and Greenspan, 2016~\cite{birenbaum2016longitudinal}}}                                 & Supervised   & Deep-learning   & 2.5D input                                                & T1-w, T1-w Gd, T2-FLAIR, PD-w                                                        & CNN         \\

Zhang et al., 2018~\cite{zhang2018ms}                                    & Supervised   & Deep-learning   & 2.5D input                                                 & T1-w, T2-FLAIR                                                                 & CNN                                                                               \\

{\color{black}{Zhang et al., 2019~\cite{zhang2019multiple}}}                                 & Supervised   & Deep-learning   & 2.5D input                                                & T1-w, T1-w Gd, T2-FLAIR                                                        & CNN         \\


{\color{black}{Ghafoorian et al., 2017~\cite{ghafoorian2017deep}}}                                 & Supervised   & Deep-learning   & 3D input                                                & T1-w, T1-w Gd, T2-FLAIR                                                        & CNN         \\

Valverde et al., 2017~\cite{valverde2017improving}, 2019~\cite{valverde2019one}                        & Supervised   & Deep-learning   & 3D input                                                & T1-w, T2-w, T2-FLAIR                                                           & CNN                                                                                              \\

Brosch et al., 2016~\cite{brosch2016deep}                                & Supervised   & Deep-learning   & 3D input                                                & T1-w, T2-w, T2-FLAIR                                                           & CNN                                                                               \\

McKinley et al., 2016~\cite{mckinley2016nabla}                           & Supervised   & Deep-learning   & 3D input                                                & T2-FLAIR                                                                       & CNN                                                                               \\

{\color{black}{Hashemi et al., 2018~\cite{hashemi2018asymmetric}}} & Supervised & Deep-learning & 3D input & T1-w, T2-w, T2-FLAIR, PD-w & CNN \\

Nair et al., 2019~\cite{nair2020exploring}                                 & Supervised   & Deep-learning   & 3D input                                                & T1-w, T1-w Gd, T2-FLAIR                                                        & CNN     \\

{\color{black}{Isensee et al., 2021~\cite{isensee2021nnu}}}                                 & Supervised   & Deep-learning   & 3D input                                                & T1-w, T1-w Gd, T2-FLAIR                                                        & CNN         \\

{\color{black}{Zhang et al., 2021~\cite{zhang2021geometric}}}                                 & Supervised   & Deep-learning   & 3D input                                                & T1-w, T1-w Gd, T2-FLAIR                                                        & CNN         \\ \hline

\end{tabular}
\end{table*}

Following these definitions, the state-of-the-art of MS lesion segmentation techniques are reviewed in the following sections. \textcolor{black}{The collection of the methods surveyed in this review is summarized in Table~\ref{tab:review}. Regarding the inclusion criteria, we searched performed a search on bibliographic repositories (Scopus and Google Scholar) with the keywords `multiple sclerosis' and `lesion segmentation' and only selected the papers that were relevant to the topic. From this selection, we excluded the papers that were only submitted to any of the public challenges from Table~\ref{tab:review}, but we still provided a brief description in Section~\ref{sec-evaluation}, to reduce the overlap with the challenge summary papers~\cite{styner20083d,carass2017longitudinal,commowick2018objective}. In addition, to avoid losing the generality of the manuscript, due to the number of new works being published every month, we set a cut off time of May 1st, 2021. We also refrained from included arXiv papers in pre-print form with preliminary results that had not yet gone through peer revision.}


\subsection{Statistical Lesion Segmentation Methods}

\subsubsection{Unsupervised methods}~Early MS lesion segmentation approaches were reliant upon unsupervised methods, which do not rely on manually annotated training data, but rather identify and segment lesions based solely on the voxel-wise intensity information of the input brain MRI. In the majority of described methods, whole brain tissue is segmented before the identification of MS lesions with a number of different techniques. In general, the Expectation Maximization (EM) algorithm~\cite{catanese2015automatic,beaumont2016automatic,roura2015toolbox}, and Gaussian Mixture Modeling (GMM)~\cite{doyle2016automatic,knight2016ms,tomas2015model} are frequently applied to classify each voxel into CSF, WM and GM clusters, and lesions are secondarily identified as outliers~\cite{weiss2013multiple,hill2015automated,roy2017effective}. In contrast, other authors have described unsupervised methods, such as the max-tree algorithm, which estimate lesions directly and obviate the need for tissue clustering~\cite{urien20163d}.

Specifically, in the method proposed by Roura et al.~\cite{roura2015toolbox}, brain tissue types were delineated using Statistical Parametric Mapping (SPM) and lesions then segmented by applying an intensity threshold on the FLAIR image. In Doyle et al.~\cite{doyle2016automatic}, brain tissues were firstly clustered using a weighted Gaussian tissue model, and the left-over outlier voxels considered as MS lesion candidates. In a similar technique described by Catanese et al.~\cite{catanese2015automatic}, the EM algorithm was applied to cluster candidate voxels into CSF, WM, and GM. Next, a graph-cut technique was applied to detect the lesions as outliers of the segmented tissues. In 2016, Knight et al.~\cite{knight2016ms} combined fuzzy clustering with an edge-based model on the original input. Thresholding and a false-positive reduction method were then applied to obtain the final lesion segmentation prediction.

\subsubsection{Supervised methods}~\textcolor{black}{Supervised  methods  aim  to training  statistical  models  that segment lesions from engineered  features  in  the annotated data.}
\textcolor{black}{Common features include spatial and intensity information. For example, features can be defined as non-local relationships and symmetry~\cite{geremia2013spatially} or similarity to other image regions or patches~\cite{guizard2015rotation}.} Various statistical learning models are applied to generate the segmentation, including logistic regression~\cite{valcarcel2018mimosa}, the k-nearest neighbor (kNN) classifier~\cite{steenwijk2013accurate}, a random forest (RF) classifier~\cite{geremia2013spatially,maier2015ms,jesson2015hierarchical}, decision trees~\cite{jog2015multi}), a multi-layer perceptron (MLP) classifier~\cite{mahbod2016automatic,santos2016evaluation}, or a Markov random fields (MRF)~\cite{doyle2017lesion} approach.

For example, Steenwijk et al.~\cite{steenwijk2013accurate} proposed to use a set of probabilistic atlases as priors for GM, WM, and CSF tissues, called tissue type priors (TTPs). These TTPs are then combined with other extracted features to create a feature vector, and then train a kNN classifier to predict whether a voxel belongs to a lesion or not. Geremia et al.~\cite{geremia2013spatially} engineered symmetry, space, and intensity features and trained an RF classifier. Similarly, Maier et al.~\cite{maier2015ms} generated spatial features by pre-processing the input voxels via intensity normalization prior to the RF classifier training, followed by post-processing techniques based on the intensity and morphology information of the initial prediction. Another approach using multi-output decision trees and spatial and intensity features was presented by Jog et al.~\cite{jog2015multi}. In Mahbod et al.~\cite{mahbod2016automatic}, engineered features based on the intensity and spatial information were fed to an artificial neural network (ANN) to provide the final segmentation output. Similarly, Santos et al.~\cite{santos2016evaluation} proposed to use a simple MLP classifier with one hidden layer to obtain MS lesion segmentation. In Doyle et al., segmentation was handled with an Iterative Multilevel Probabilistic Graphical model (IMaGe) with two MRF blocks at the region level ~\cite{doyle2017lesion} based on tissue class priors and local and neighboring information.

\subsection{Deep-learning based Lesion Segmentation Methods}

\textcolor{black}{Compared to classical statistical methods, deep-learning methods can generate representative features automatically during training using specific layers such as convolutional blocks from 2D slices, either from the same view~\cite{aslani2019multi} or combining slices from different views to generate a so-called 2.5D image~\cite{birenbaum2016longitudinal,zhang2019multiple}, or 3D volumes~\cite{hashemi2018asymmetric,zhang2021geometric}}. By processing a complex representation in a high-dimensional feature space, these methods can achieve better performance. 

One of the first deep learning approaches to segment MS lesions was presented by Ghafoorian et al. with a five-layer CNN network~\cite{ghafoorian2017deep}. At that point in time, CNN architectures were mostly used for image classification. Gafhoorian’s method essentially used this approach to obtain a voxel-level segmentation by classifying a patch for each possible image voxel. In the same year, another 3D network was developed by Vaidya et al. with four layers for multi-channel 3D patches~\cite{vaidya2015longitudinal}. Similarly, Valverde et al. also extended this approach to 3D patches to include 3D information from the processed MRI data to predict the label of the center voxel~\cite{valverde2016multiple, valverde2017improving}. The authors also proposed a cascaded architecture to provide high sensitivity in the first network, with the trade-off of an increased number of false positives; and a second network to counter the effect of false positives. While approaches based on the classification of whole patches for single pixels provide a solution for segmentation, they are inefficient and require a large amount of processing time and resources. If the average clinical MRI contains millions of voxels, using such an approach would require millions of inferences of adjacent patches to segment the whole brain.

{\color{black}For a more efficient inference, fully convolutional networks (FCN) bypass the use of fully connected layers for the output and instead use only convolutional layers to provide a pixel-wise prediction. One of the first fully convolutional CNNs was presented by Havaei et al.~\cite{havaei2016hemis} to solve the problem of missing modalities.} In their segmentation methodology, multi-channel 2D images were fed as input into a multi-stage convolutional pathway to deal with subjects that had different numbers and types of input images. The proposed network contains a front-end block, an abstraction block, and a back-end block. The front-end block uses independent convolutional blocks to yield information mapping of different modalities, and the total number of input modalities is flexible per case. The abstraction layer then interprets the statistical information of the input mapping; finally, the front-end block, with two convolutional layers, generates a pixel-wise probability classification output. In order to benefit from multichannel information, Roy et al. developed a CNN structure with multiple convolutional pathways for 2D patches of two independent input modalities (FLAIR, T1-w imaging)~\cite{roy2018multiple}. The output of these two pathways is concatenated and a final convolutional pathway used to yield the lesion segmentation.

{\color{black}More recent FCN architectures have favored encoder-decoder architectures, popularized by the U-Net architecture~\cite{ronneberger2015u,cciccek20163d},} to reduce network complexity while maintaining a similar performance, or in some instances, better performance. Brosch et al.~\cite{brosch2016deep} developed a deep 3D convolutional seven-layer encoder network with shortcut connections (CENs). The authors defined a mapping function to transfer multi-channel MRI to probabilistic lesion masks. High-level representations of the input were abstracted using convolutional layers, along with pooling layers to reduce the spatial size of the representation and the number of parameters for training.

In related work, McKinley et al. proposed `Nabla-net’, a FCN model for MS lesion segmentation~\cite{mckinley2016nabla} that also adopted an encoder-decoder structure to extract higher-level representations of the input data with 18 layers in total. The authors subsequently described `DeepSCAN’~\cite{mckinley2020automatic}, another architecture that uses 3D input patches and converts them into 2D slices on the encoding path. Then, rather than recovering the initial 3D shape, segmentation is performed on 2D slices to obtain the final output. This combination of 3D and 2D processing maintains part of 3D information while simplifying the final segmentation steps. The method also involves an uncertainty-based loss function using a combination of focal loss and label-flip loss to account for noisy labels. Finally, the same group extended their approach to simultaneously segment brain structures and lesions~\cite{mckinley2021simultaneous}. Another example of an encoder-decoder architecture for MS lesion segmentation and uncertainty prediction was described by Nair et al.~\cite{nair2020exploring}. Their work explored four voxel-based uncertainty estimate measures using the Monte Carlo (MC) dropout technique presented by Gal and Ghahramani~\cite{gal2016dropout}. Gal and Ghahramani showed that minimization of the Kullback-Leibler (KL) divergence, which captures the uncertainty in the model, is equivalent to the minimization of cross-entropy loss of a network with dropout applied. The uncertainty measures explored by Nair et al. included the predicted variance learned from the training data and three stochastic sampling-based measures including predictive entropy, the variance of the MC samples, and mutual information. Using these uncertainty measures, they also improved voxel-wise true positive and false-positive rates, which can benefit the identification of small lesions.

Rather than using normal convolutions to represent local relationships between pixels, Andermatt et al. implemented an encoder-decoder architecture based on multidimensional gated recurrent units (MD-GRU)~\cite{andermatt2017automated}. A gated recurrent unit (GRU)~\cite{chung2014empirical} is a feature extraction module based on convolution and a gating mechanism, working as a building block for recurrent neural networks (RNN). These GRU layers are capable of encoding neighbourhood relationships without the constraint of the traditional convolution kernel size. The authors also reported that this modification allowed the model to converge faster.

{\color{black}As a culmination of the U-Net architecture, Isensee et al. developed the nnU-Net segmentation framework for general medical image segmentation~\cite{isensee2021nnu}. This framework dynamically adapts the encoder-decoder hyperparameters to the specifics of any given training dataset. The essence of this approach is the in-depth design of adaptive preprocessing, training schemes, and inference. All design choices required to adapt to new segmentation tasks are made in a fully automated manner, avoiding the need for manual interaction. For each task, n-fold cross-validations are systematically performed for different U-Net models; and the mean foreground DSC score employed as the criterion to choose and submit a final model. This framework was tested on the MS lesion segmentation task, outperforming the 3D-to-2D (DeepSCAN) CNN architecture.}

{\color{black}
Finally, while encoder-decoder architectures have become a de facto standard, in~\cite{aslani2019multi}, Aslani et al. designed a fully convolutional CNN framework to process the features of different modality 2D slices separately, without an explicit decoder based on the ResNet architecture for natural imaging~\cite{kaiming2016resnet}. Specifically, the input 2D slices for each modality are passed through a modality-specific feature extractor. Next, the features from different modalities and resolutions are fused via a multi-modal feature fusion block and multi-scale upsampling mechanisms to obtain the final lesion segmentation predictions.
}

\section{Potential clinical lesion segmentation applications\label{sec-les-app}}

In Section~\ref{sec-les-seg}, we provided a detailed survey of the reviewed cross-sectional lesion segmentation methods. In practical clinical applications, the analysis of longitudinal lesion evolution is an important measure of disease progression and treatment efficacy that informs optimal management of individual patients. Additionally, the clinical adoption of fully automated segmentation methods is hampered by domain bias imparted by different MRI acquisition protocols, in which a specific model may underperform due to inter-scanner/protocol variances between the training and testing dataset. In this section, we will further discuss technical approaches to address these key translational requirements.

\subsection{Lesion Activity Analysis}
To assess the lesion evolution over time, longitudinal analysis of a series of MRI images is required. Methods that process each time point independently are less reliable due to differences in the images that can cause discrepant segmentation masks when temporal information and redundant features (such as stable image regions) are not considered. Most methods apply subtraction as a simple technique to remove “stable” disease and highlight temporal changes. Some unsupervised strategies apply thresholding to these subtraction images to classify new lesions, such as the works of Ganiler et al.~\cite{ganiler2014subtraction} and Cabezas et al.~\cite{cabezas2016improved}.

Jain et al. presented a subtraction-based framework called MSmetrix-long, using the lesion segmentation from two time points and subtraction imaging ~\cite{jain2016two}. Their framework firstly applied MSmetrix-cross~\cite{jain2015automatic} to get a cross-sectional lesion segmentation of each time point separately. Then, the FLAIR-based subtraction images were created for the two directions via co-registration and bias normalization. Similarly, Schmidt et al.~\cite{schmidt2019long} presented an approach to detect lesion changes using cross-sectional masks of each timepoint~\cite{schmidt2012lst} and their FLAIR intensity differences. Using the intensity differences in the normal appearing white matter region, they defined a rule to detect significant intensity differences for the lesion areas.

Using statistical modeling, Cerri et al.~\cite{cerri2020longitudinal} proposed a methodology capable of segmenting multiple time points at the same time. Extending their cross-sectional approach~\cite{cerri2021contrast} where each input image is processed separately, they exploited the common spatial features shared by all the time points from the same subject. Subject-specific latent variables were introduced to the segmentation priors and likelihood, linking different time points with a statistical dependency.

The first CNN architecture for new T2-w lesion segmentation in longitudinal brain MR images was proposed by Salem et al.~\cite{salem2020fully}. By using an unsupervised registration module to obtain the deformation field between timepoints with a supervised module for lesion segmentation, the overall method can be trained in an end-to-end fashion. In the same year, Gessert et al.~\cite{gessert2020multiple} applied a CNN model for lesion activity segmentation using an attention mechanism~\cite{vaswani2017attention}. Specifically, the authors engineered an architecture with an independent encoding pathway for each timepoint. Then, the pathways are fused with the bottleneck of the network and the last few layers of the CNN concatenated to the fusion features to generate the final segmentation. As an extension of this work, Gessert et al. described another fully 4D architecture to segment lesions over multiple time points~\cite{gessert20204d}, proposing a 3D ResNet-based multi-encoder-decoder CNN architecture in which, assuming the  number of time points is T, a T-path encoder with shared weights is used to process all the input time points in parallel. Temporal aggregation is then completed via convolutional gated recurrent units (convGRUs), which are also employed as long-range connections from the encoders to decoders.

\subsection{Domain Adaptation}

The intensity contrast of brain tissues and MS lesions varies from the scanner to scanner, and even within the same scanner depending on the choice of acquisition  parameters. Therefore, MS lesion segmentation methods in clinical practice can face several issues when using images from different domains due to the so-called domain shift issue. Multi-source domain adaptation and harmonization techniques aim to adapt a model to data acquired from different sources. This process can be achieved in a supervised manner when labels of the new domain are available; or in a completely unsupervised manner, when only the source of the new domain is known.

Ghafoorian et al.~\cite{ghafoorian2017transfer} studied the effectiveness of fine-tuning on a fitted model with additional annotated data from a new target domain. Experiments were conducted on a CNN, where the parameters of the shallowest $ith$ layers were frozen and the remaining $d-i$ layers were fine-tuned with dataset D from another domain. In addition, the optimal number of frozen layers and the smallest size of the fine-tuning dataset D for a proper adaptation were also investigated.


Based on the assumption that FC layers are more sensitive to domain bias than CNN layers, Valverde et al.~\cite{valverde2019one} proposed a few-shot supervised technique by re-training the FC layers of the source model with target images to tackle domain adaptation. In general, encoded knowledge in the source model can be effectively adapted to an unseen target intensity domain. This is because convolutional layers contain related features that can be transferred to unseen data while only re-training the FC layers. In that sense, domain adaptation is performed by retraining all or some of the source FC layers using images from the target domain. Re-use of implicit knowledge trained on the source model significantly lowers the number of weights required to optimize on the target model, reducing the requisite number of training images and avoiding over-fitting of the model.

For supervised domain adaptation, sub-structure re-training is another effective approach in addition to the adversarial learning strategy, given access to the labels for both source and target domains. In contrast, generative adversarial networks (GANs) have been applied to alleviate problems that arise in the context of limited amounts of training data in unsupervised settings. Zhang et al. presented MS-GAN, an MS lesion segmentation tool~\cite{zhang2018ms} in which a generator network (G), based on the multi-channel encoder-decoder structure, is combined with multiple discriminator networks (D) for different input channels. By taking advantage of adversarial training, the authors achieved precision lesion segmentation without the need for experimental parameter tuning or other pre- or post-processing steps.

Kamnitsas et al. investigated unsupervised domain adaptation using adversarial neural networks to counter the effects of domain shift~\cite{kamnitsas2017unsupervised}. For the test domain, this method does not require additional annotation, and the generator is a fully convolutional neural network (FCN) for the segmentation task. A domain discriminator is implemented as a second neural network that distinguishes between images from the source and target domains. The training of the discriminator is exposed to the generator and vice versa. In that manner, the generator tries to increase the domain classification loss, thus helping to extract fewer domain-specific features and making the segmentation less sensitive to domain change.

Similarly, Palladino et al.~\cite{palladino2020unsupervised} explored cycle-consistent adversarial networks (CycleGAN) to map the distribution of the source domain to the target domain and a 3D U-Net trained in the target domain to segment the style-transferred images. CycleGAN~\cite{zhu2017unpaired} is a type of GAN architecture proposed for style-transfer commonly used with unpaired data (i.e. images from the same subject in different domains). The generator is designed to find a mapping function that makes the source domain-derived images indistinguishable from real target images. Furthermore, the network has an inverse mapping, creating a cycle consistency loss for the entire framework. Both generator and discriminator follow CNN-based encoder-decoder architectures.  

In addition, Ackaouy et al.~\cite{ackaouy2020unsupervised} proposed a medical image segmentation framework called Seg-JDOT that uses unsupervised optimal transport for domain adaption. This framework ensures similar inference results for a model within different domains sharing similar representations. Tested on MS segmentation task with a 3D U-Net architecture, the model yielded promising performance in the target domain over standard training.

Finally, Dewey et al.~\cite{dewey2019deepharmony} explored a different path by directly harmonizing the contrast of the input images among different domains with an architecture called DeepHarmony. The model is based on the U-Net architecture with additional concatenation steps between input contrasts and the final feature map, only altering the input contrasts and not creating entirely new contrasts. The test set included a small overlap cohort (n = 8) from two different acquisition protocols. The evaluation on this set showed a reduced inconsistency between the generated images with different inputs.


\section{Evaluation\label{sec-evaluation}}

The evaluation and comparison of different models proved to be difficult since not all of the reviewed methods made their datasets and trained models publicly available for evaluation. In this survey, we have grouped the methods that used a common public dataset for evaluation.

\subsection{Public Datasets}


\textcolor{black}{The first MICCAI MS lesion segmentation challenge (MICCAI-MSSEG 2008) was held in 2008~\cite{styner20083d}. The imaging data provided for this challenge was acquired separately from Children's Hospital Boston (CHB) and University of North Carolina (UNC). The scans collected by UNC had slice thickness of 1mm and in-plane resolution of 0.5mm and they were all acquired on Siemens 3T Allegra MRI scanner. Preprocessing steps included rigidly registering each subject to a common reference frame and reslicing all images to an isotropic voxel spacing to ease the segmentation and evaluation process. 10 CHB and 10 UNC cases were provided with manual segmentations from CHB expert for training; 15 CHB and 10 UNC cases were provided as testing cases without labels.}

The ISBI 2015 MS lesion challenge focused on longitudinal lesion segmentation task~\cite{carass2017longitudinal}. All scans were acquired using the same protocol on a 3.0 Tesla MRI scanner. Training data provided by this challenge was derived from 5 subjects with a consecutively acquired brain MRI scans (mean of 4.4 time-points). The challenge also provided another two test sets from 10 (Test Set A) and 4 (Test Set B) subjects, respectively. Consecutive timepoints were acquired with a one-year gap. All images were annotated by two human experts, yielding a consensus delineation. This challenge additionally provided preprocessed images, following bias correction using N4 algorithm, skull and dura stripping, and rigid registration to 1mm isotropic MNI template. The training dataset and Test Set A were released to participants before the challenge with the corresponding manually labelled MS lesion segmentation masks. 

\textcolor{black}{Another MICCAI MS lesion segmentation challenge (MICCAI-MSSEG-1 2016) was held in 2016}, providing a high-quality database of 53 MS cases acquired using different scanners following the Observatoire français de la sclérose en plaques (OFSEP) protocol recommendations~\cite{commowick2018objective}. For each MS patient scan, seven expert manual annotations were gathered, and a consensus ground truth was provided. The challenge organizers provided raw images and preprocessed images. The preprocessing steps included denoising with a non-local means algorithm, rigid registration of each modality to the FLAIR image, skull-stripping using volBrain based on T1 image, and bias correction using the N4 algorithm.

\textcolor{black}{Recently, a complementary challenge to MICCAI-MSSEG-1 named MICCAI-MSSEG-2 was held in 2021 and focused on lesion activity, rather than cross-sectional lesion segmentation. This dataset comprised 100 MS patients with FLAIR MRI scans from two different time points (with a 1 to 3 years interval between the time points). The data was collected from the same OFSEP cohort, a French MS registry from different centers and scanners. Rigid registration was applied to warp and align the baseline and follow-up images into a middle space. No further preprocessing steps were applied and the use of additional steps was left to the participants' discretion. Similarly to MICCAI-MSSEG-1, all subjects were annotated by four different experts and a global consensus of the different masks was used as the final evaluation ground truth.}

\subsection{Evaluation Metrics}

The most common performance evaluation metrics,  such  as DSC and True positive rate (TPR), are listed in Table~\ref{tab:metrics}. We chose these metrics based on their extended use on the vast majority of papers, and their application on the different public challenges focusing on MS lesion segmentation.

\textcolor{black}{
The MICCAI-MSSEG 2008 challenge used a set of different evaluation metrics. For this review, we decided to only include lesion-wise metrics, including lesion-wise TPR (LTPR), and lesion-wise FPR (LFPR). While distance metrics and the absolute volume difference were also provided during the evaluation, both metrics are hard to interpret (a total distance value is given for multiple focal lesions) or misleading (volume difference does not account for and have been rarely used since then). While ISBI 2015 and MICCAI-MSSEG-1 2016 also adopted these metrics, they extended the evaluation to voxel-wise overlap metrics, such as the DSC and the positive predictive value (PPV). Finally, MICCAI-MSSEG-2 focused on new lesions and employed the F1 and Dice score for the detection and segmentation performance respectively. In addition, the number of falsely detection lesions and their volume in stable subjects was also calculated to evaluate the performance in subjects without new lesions.}

\textcolor{black}{In summary, the most commonly employed metrics to evaluate the performance of MS lesion segmentation methods are included in Table~\ref{tab:metrics}. 
}

\begin{table}[t]
\centering
  \caption{Evaluation metrics commonly used for MS segmentation in all the challenges, where TP, FN and FP stand for True Positives, False Negatives and False Positives, respectively.}
  \label{tab:metrics}
    \begin{tabular}{ll}
    \hline
Metric Name                                                               & Equation                                                                                                                                                                                               
\\ \hline \\

\begin{tabular}[c]{@{}l@{}}Dice similarity\\ coefficient (DSC)\end{tabular}     & \begin{math}\frac{2TP}{2TP+FN+FP}\end{math} \\ \\

True positive rate (TPR)  & \begin{math}\frac{TP}{TP+FN}\end{math} \\ \\

False positive rate (FPR) & \begin{math}\frac{FP}{TP+FN}\end{math} \\ \\

Positive predictive value (PPV) & \begin{math}1-\frac{FP}{TP+FN}\end{math} \\ \\

F1 Score & \begin{math}\frac{TP}{TP+\frac{1}{2} (FP+FN)}\end{math} \\ \\
\end{tabular}
\end{table}

\begin{table}[!htb]
  \centering
  \caption{\textcolor{black}{Review of MS segmentation methods using MICCAI-MSSEG 2008 challenge dataset}}
  \label{tab:miccai2008results}
\scriptsize
\begin{tabular}{llllll}
\hline\hline

Method & Score  & LTPR & LFPR \\ \hline
Bricq et al., 2008~\cite{bricq2008ms} & 82.12 & 0.44$\pm$0.17 & 0.57$\pm$0.24\\ 
Souplet et al., 2008~\cite{souplet2008auto} & 80.00 & 0.54$\pm$0.20 & 0.73$\pm$0.18 \\
Shiee et al., 2008~\cite{shiee2008multiple} & 79.90 & 0.52$\pm$0.24 & 0.73$\pm$0.22 \\ \hline

Valverde et al., 2017~\cite{valverde2017improving} & 87.12 & 0.62$\pm$0.18 & 0.46$\pm$0.22 \\
Jesson et al., 2015~\cite{jesson2015hierarchical} & 86.94 & 0.49$\pm$0.22 & 0.28$\pm$0.24 \\ 
Guizard et al., 2015~\cite{guizard2015rotation} & 86.11 & 0.50$\pm$0.20 & 0.43$\pm$0.22 \\
Tomas-Fernandez et al., 2015~\cite{tomas2015model} & 84.46 & 0.47$\pm$0.18 & 0.45$\pm$0.23 \\
Brosch et al., 2016~\cite{brosch2016deep} & 84.07 & 0.52$\pm$0.20 & 0.51$\pm$0.19 \\
Roura et al., 2015~\cite{roura2015toolbox} & 82.34 & 0.50$\pm$0.24 & 0.42$\pm$0.23 \\
Geremia at al., \cite{geremia2011spatial} & 82.07 & 0.55$\pm$0.20 & 0.74$\pm$0.13 \\
Zhang et al., 2019~\cite{zhang2019multiple} & 83.28 & 0.61$\pm$0.29 & 0.40$\pm$0.31 \\
Cerri et al., 2021~\cite{cerri2021contrast} & 76.91 & 0.38$\pm$0.22 & 0.61$\pm$0.27 \\
Zhang et al., 2018~\cite{zhang2018ms} & 73.10 & 0.78$\pm$0.15 & 0.72$\pm$0.15 \\

\end{tabular}
\end{table}

\begin{table}[!htb]
  \centering
  \caption{\textcolor{black}{Review of MS segmentation methods using ISBI 2015 challenge dataset}}
  \label{tab:isbi2015results}
\tiny
\begin{tabular}{llllll}
\hline\hline
Method & Score  & Dice & LTPR & LFPR \\ \hline
Jesson et al., 2015~\cite{jesson2015hierarchical} & 90.70 & 0.58$\pm$0.12 & 0.13$\pm$0.16 & 0.20$\pm$0.12 \\
Maier et al., 2015~\cite{maier2015ms} & 90.28 & 0.61$\pm$0.17 & 0.37$\pm$0.20 & 0.27$\pm$0.24 \\
Catanese et al., 2015~\cite{catanese2015automatic} & 89.81 & 0.60$\pm$0.13 & 0.41$\pm$0.21 & 0.31$\pm$0.17 \\ \hline

Zhang et al., 2019~\cite{zhang2019multiple} & 93.12 & 0.68$\pm$0.14 & 0.58$\pm$0.23 & 0.19$\pm$0.15 \\
Isensse et al., 2021~\cite{isensee2021nnu} & 92.87 & 0.68$\pm$0.14 & 0.52$\pm$0.23 & 0.16$\pm$0.13 \\
Zhang et al., 2021~\cite{zhang2021geometric} & 92.73 & 0.64$\pm$0.16 & 0.48$\pm$0.23 & 0.13$\pm$0.12 \\
Hashemi et al., 2018~\cite{hashemi2018asymmetric} & 92.49 & 0.58$\pm$0.16 & 0.41$\pm$0.21 & 0.09$\pm$0.10 \\
Aslani et al., 2019~\cite{aslani2019multi} & 92.12 & 0.61$\pm$0.15 & 0.41$\pm$0.21 & 0.20$\pm$0.14 \\
Andermatt et al., 2017~\cite{andermatt2017automated} & 92.08 & 0.63$\pm$0.14 & 0.49$\pm$0.22 & 0.20$\pm$0.14 \\
Valverde et al., 2019~\cite{valverde2019one} & 91.33 & 0.63$\pm$0.13 & 0.37$\pm$0.19 & 0.15$\pm$0.15 \\
Birenbaum and Greenspan, 2016~\cite{birenbaum2016longitudinal} & 90.07 & 0.63$\pm$0.14 & 0.57$\pm$0.24 & 0.50$\pm$0.18 \\
Ghafoorian et al., 2017~\cite{ghafoorian2017deep} & 87.72 & 0.57$\pm$0.13 & 0.35$\pm$0.18 & 0.47$\pm$0.18 \\
Valcarcel et al., 2018~\cite{valcarcel2018mimosa} & 86.92 & 0.50$\pm$0.14 & 0.43$\pm$0.21 & 0.58$\pm$0.22 \\

\end{tabular}
\end{table}

\subsection{Comparisons of Reviewed Methods}
\textcolor{black}{For MICCAI-MSSEG 2008 challenge and ISBI 2015 challenge, the evaluation of the top-3 participating methods and the post-challenge published submissions are summarized in Table~\ref{tab:miccai2008results} and Table~\ref{tab:isbi2015results}, respectively. Since the MICCAI-MSSEG 2008 challenge did not evaluate the DSC metric, DSC was only reported for the ISBI 2015 submissions. Finally, the testing set results for MICCAI-MSSEG-1 2016 were evaluated by the organizers using Docker containers of the submitted methods and summarized on the corresponding paper~\cite{commowick2018objective}. Moreover, there the majority of papers we reviewed did not use the training data from the challenge. Therefore, we decided not to include this challenge in the corresponding evaluation results. }


\begin{figure*}[!htb]
\centering
    \begin{subfigure}[b]{0.3\textwidth}
    \centering
    \includegraphics[width=\textwidth]{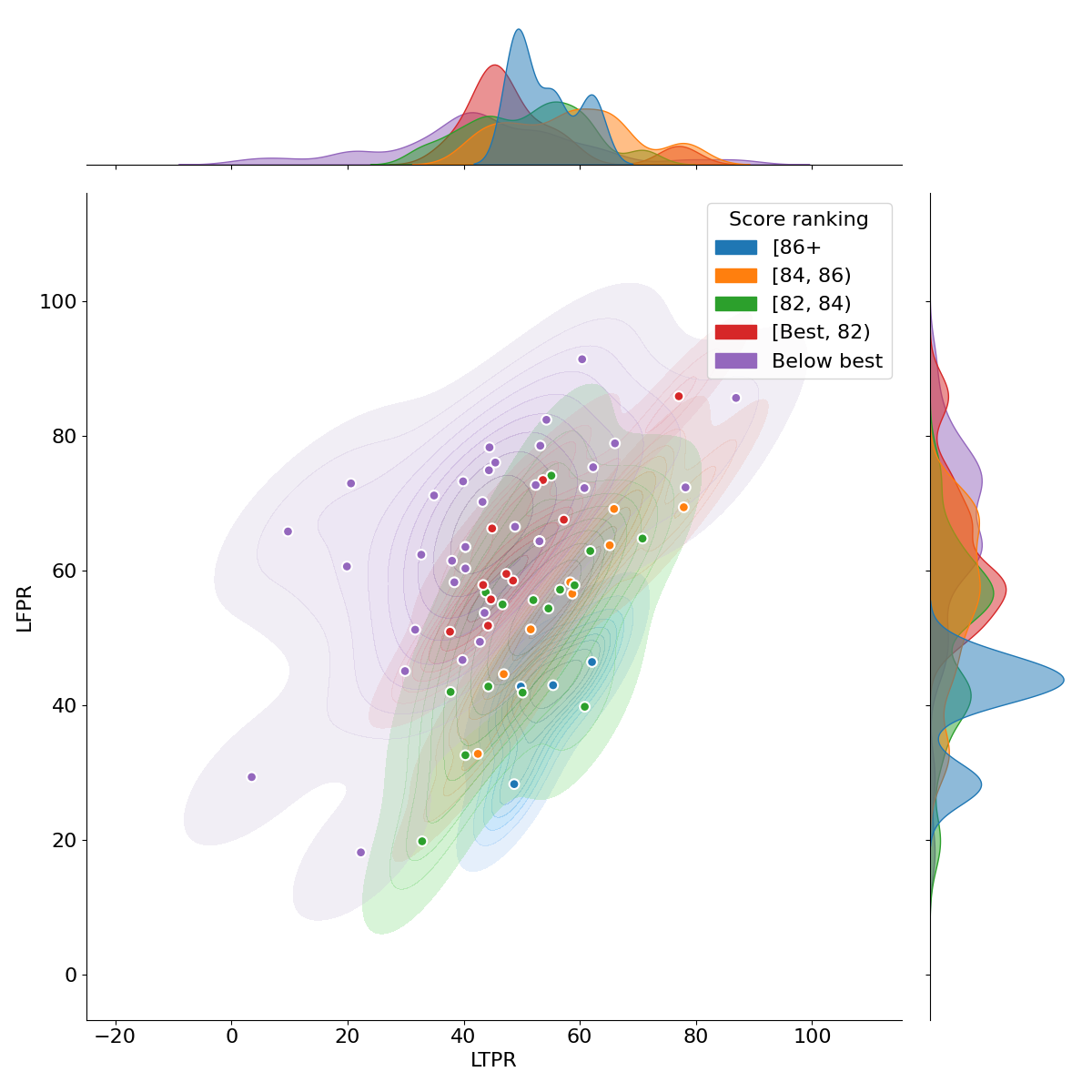}
    \caption{LTPR vs. LFPR (MICCAI-MSSEG 2008)}
    \end{subfigure} 
    \begin{subfigure}[b]{0.3\textwidth}
    \centering
    \includegraphics[width=\textwidth]{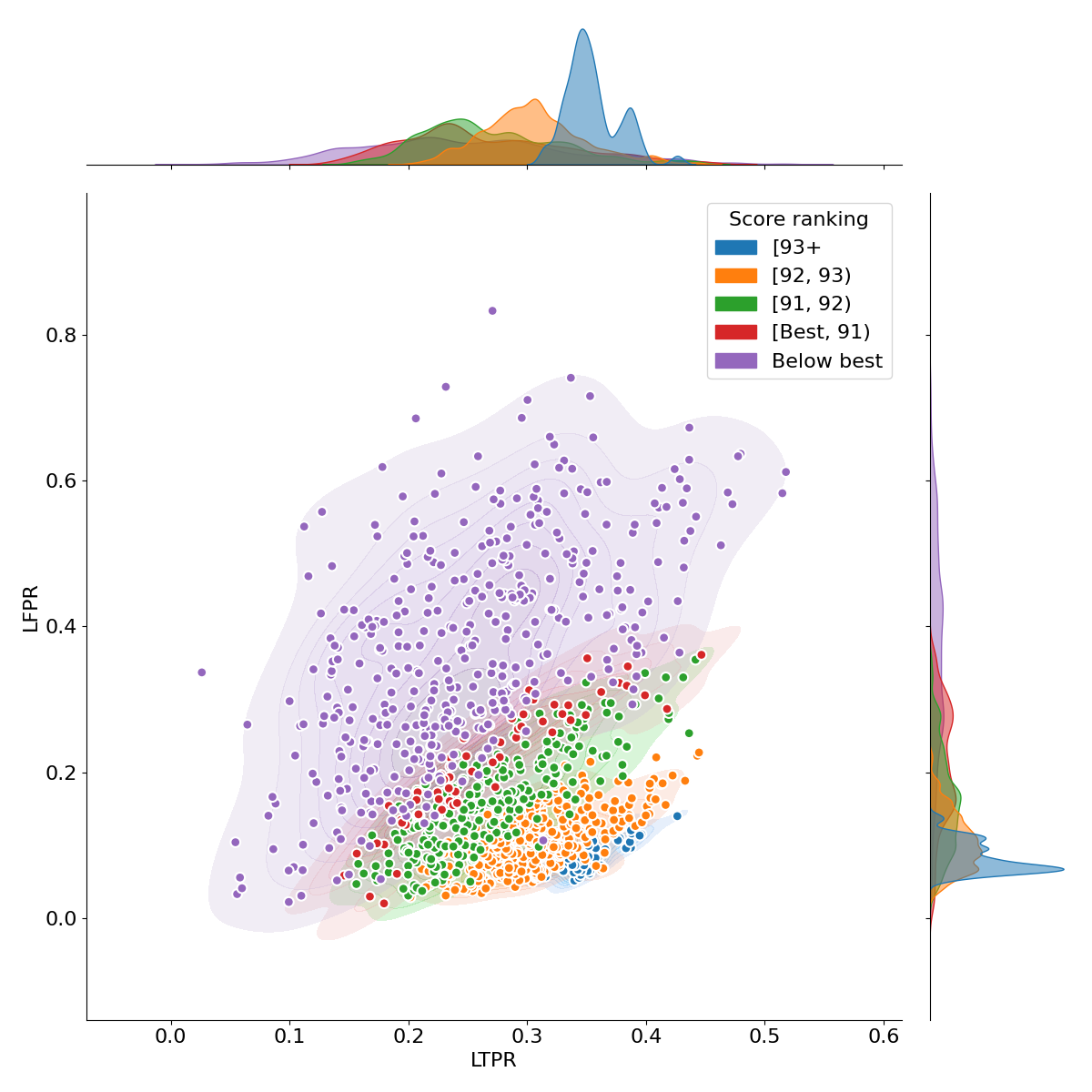}
    \caption{LTPR vs. LFPR (ISBI 2015)}
    \end{subfigure} 
    \begin{subfigure}[b]{0.3\textwidth}
    \centering
    \includegraphics[width=\textwidth]{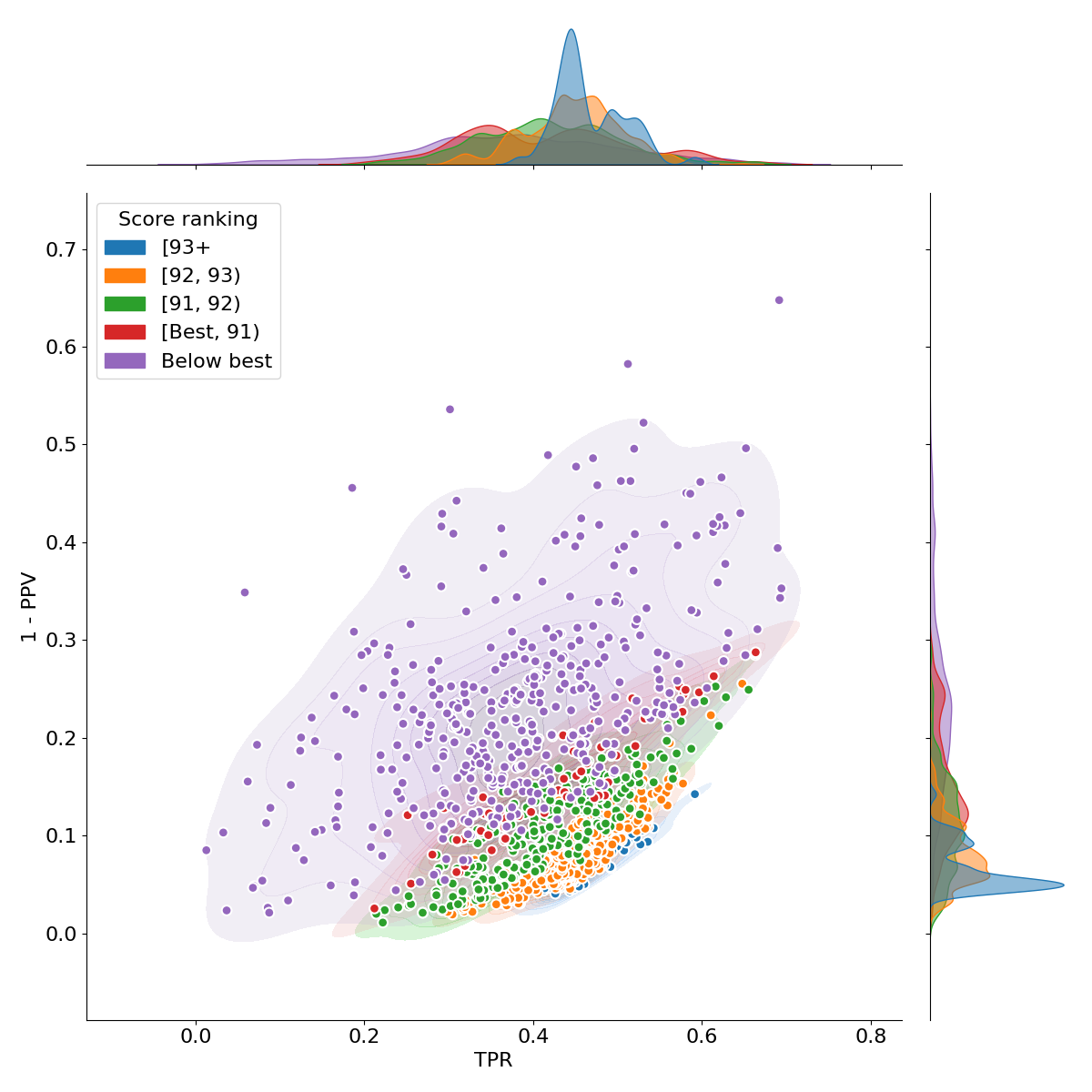}
    \caption{TPR vs. 1-PPV (ISBI 2015)}
    \end{subfigure}
\caption{\textcolor{black}{Lesion-wise True Positive Rate (LTPR) vs. Lesion-wise False Positive Rate (LFPR) for (a) MICCAI-MSSEG 2008 and (b) ISBI 2015 submissions; (c) True Positive Rate (TPR) vs. 1-Positive Predictive Value (which is identical to FPR) for ISBI 2015 submissions. The submission with the best score participated in the corresponding challenge has been set as the baseline and marked in purple.}}
\label{fig:challenge_results}
\end{figure*}

\begin{figure*}[!htb]
\centering
    \begin{subfigure}[b]{0.4\textwidth}
    \centering
    \includegraphics[width=\textwidth]{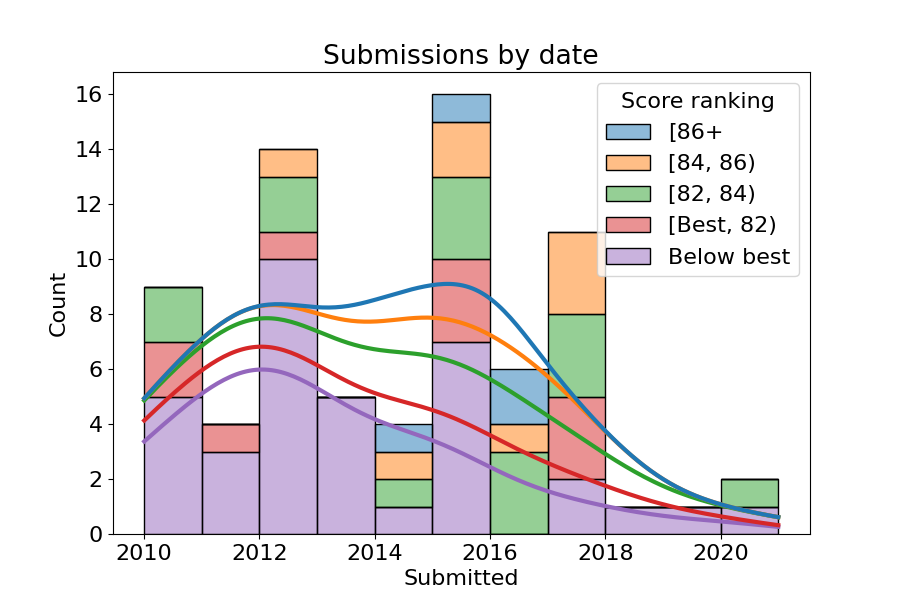}
    \caption{MICCAI-MSSEG 2008}
    \end{subfigure} 
\hspace{1.5cm}
    \begin{subfigure}[b]{0.4\textwidth}
    \centering
    \includegraphics[width=\textwidth]{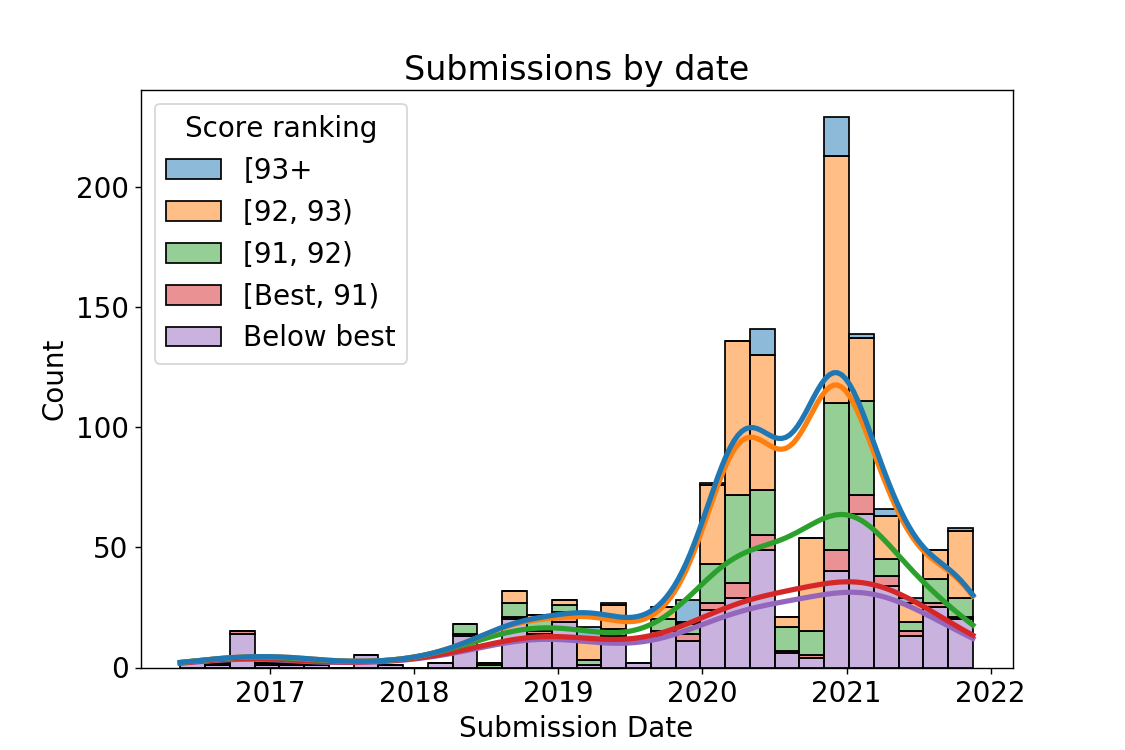}
    \caption{ISBI 2015}
    \end{subfigure}
\caption{\textcolor{black}{Performance evolution over time of (a) MICCAI-MSSEG 2008 and (b) ISBI 2015 challenge. The submission with the best score participated in the corresponding challenge has been set as the baseline and marked in purple. Note that the time stamps for all the methods presented at the MICCAI-MSSEG 2008 challenge were set to the time that the post-challenge submission site was reopen in 2010.}}
\label{fig:challenge_scores}
\end{figure*}


In general, supervised methods with 3D inputs, such as described by McKinley et al.~\cite{mckinley2016nabla} and Valverde et al.~\cite{valverde2016multiple}, outperformed other strategies. In contrast, unsupervised methods and methods using 2D image sequences seemed to have a lower performance~\cite{danelakis2018survey}.

In terms of supervised classification, the performance was more robust when a deep learning model or random forest was applied. Deep learning currently dominates the field, with the majority of recent works on lesion segmentation focusing on CNN-based models, and specifically U-Net architectures. Deep learning methods have been applied to both the lesion activity segmentation task and the multi-source domain adaptation task with promising results that approximate the performance of the average human expert rater.

However, the performance and accuracy of a particular method are insufficient to determine clinical applicability. For example, computational efficiency is also important in practice; and deep learning models are constrained by their computational complexity (time, memory, and power). However, with the implementation of GPU resources for training and inference, the time cost associated with deep learning methods can be significantly reduced, at the expense of an increased carbon footprint and power consumption.

\subsection{Meta-analysis of the Post-challenge Submissions}
\textcolor{black}{
Since the challenges were held, the MICCAI-MSSEG 2008 and ISBI 2015 website platforms have remained open and there have been several submissions, including the works reviewed in this survey and other unknown teams. We thus performed a meta-analysis of these two challenges to illustrate the evolution of the performance of MS lesion segmentation techniques through time with the datasets as a proxy.}

\textcolor{black}{True and false positive detections and segmentations, are the most basic measure to understand whether a model is overestimating or underestimating the lesion load for a cohort. As such, both challenges focused on TPR and FPR measures for lesion detections (LTPR and LFPR, respectively), and the ISBI challenge also focused on similar metrics for segmentation (TPR and PPV). To illustrate the balance between false and true predictions, Fig.~\ref{fig:challenge_results} provides scatterplots with the estimated distributions of each score range for both challenges. The submission distributions per score have a diagonal shape in the 2-dimensional plots, which highlights the importance of balancing these measures to obtain the highest score values (represented in blue). Furthermore, as the range of score gets higher the distributions get smaller (and less populated) and start to focus more on the TPR values than the FPR. This analysis, suggests that the segmentation methods are reaching a ceiling in terms of accurate segmentation.}

\textcolor{black}{According to the 2008 challenge description~\cite{styner20083d}, a submitted method with an overall score of 90 points can be roughly considered to be comparable to human expert performance. In the case of the ISBI 2015 challenge, the score was created to mimic the original score from MSSEG 2008~\cite{carass2017longitudinal}, therefore, we decided to extend that assumption to the results of its platform. Thus, a similar conclusion about the performance evolution can be obtained from Fig.~\ref{fig:challenge_scores}. A quick glance at the overall score evolution over time shows that the most recent methods have struggled to obtain the best scores in the past submissions for the 2008 challenge, while a peak in over 90 scores can be observed for the ISBI one in 2021 after an increasing trend at the start of 2020. Deep learning methods started to gain attention around 2015 as the highest scores in the MSSEG 2008 challenge highlight. This year also marked the introduction of the U-net architecture~\cite{ronneberger2015u}, which has become the state-of-the-art CNN model since then~\cite{isensee2021nnu}. While this architecture supposed a revolution on the medical image domain, as evidenced by the dominance of the nnU-net technique by Isensee et al.~\cite{isensee2021nnu}, the improvements and new ideas have stagnated since then, with marginal improvements in the results (as evidenced by the large number of works in the same score range for the ISBI 2015) and numerous extensions that still rely heavily on the strengths of an encoder-decoder architecture~\cite{andermatt2017automated,valcarcel2018mimosa,hashemi2018asymmetric,zhang2018ms,zhang2019multiple,isensee2021nnu,zhang2021geometric,aslani2019multi}. This, in turn, suggests again that the current deep learning approaches are reaching the performance ceiling for these benchmarks.
}

\textcolor{black}{Finally, these two challenges represent two different aspects of lesion segmentation. On one hand, the MSSEG 2008 challenge offers a glance at the generalization of the segmentation methods by combining acquisitions from two different centers. As such, the ceiling is probably lower (as suggested by the lower overall scores that failed to reach 90) due to the issues introduced by domain shifts in the data. On the other hand, ISBI 2015 focuses on the reproducibility of segmentation results in a single center study of longitudinal data. The ceiling, therefore, should be higher (as exemplified by the higher overall scores surpassing the 90 threshold). Furthermore, due to the nature of the study (all the subjects have multiple points that are fairly similar), good performance on a subject, should be magnified when analysing all its timepoints. Consequently, it is not unexpected that the MSSEG 2008 submissions reached their peak before the ISBI 2015 challenge. Besides, newer methods tend to favor evaluation with newer benchmarks, and at the time of writing, the MSSEG 2008 challenge is more than a decade old.}

\subsection{Analysis of MICCAI-MSSEG-2}
\textcolor{black}{The majority of the participants used a deep learning approach, most of them based on the U-net architecture, while only one of the teams focused on subtraction imaging and classical statistical methods for their submission~\cite{msseg2}. For the evaluation, the organisers provided multiple rankings (one for each metric) and no overall score was used to rank all methods in a summarized manner, similarly to the MSSEG-1 2016 challenge~\footnote{The analysis of the results is based on the evaluation slides at~\url{https://files.inria.fr/empenn/msseg-2/Challenge_Day_MSSEG2_Results_2021.pdf}}. Since these metrics were measured from the consensus of 4 experts, raters were also ranked and could be compared to the predictions of each team. For the lesion number and volume in stable cases (those without new lesions), four of the submitted works obtained results that were superior or comparable to human experts (teams Lyle, SCAN~\cite{mckinley2020automatic}, Neuropoly-2 and PVG). This, however, it is not surprising in itself. Methods that had difficulties to find new lesions (those with a low TPR), would also struggle to find new lesions on stable patients as a consequence. In fact, these methods were not able to reach the standard of human experts for the F1 and Dice score on the subjects with new lesions.}

\textcolor{black}{The organisers also observed variations on the ranking of all the methods depending on the metric. In fact, the top 5 methods were fairly different for each ranking. Furthermore, one of the experts (number 4), was outperformed by different submissions on the Dice (teams MedICL, LaBRI-IQDA, SNAC, LaBRI-D\&E and NVAUTO) and F1 score (teams Mediaire-B, Empenn and Mediaire-A). Moreover, while CNN approaches were capable of outperforming a less experienced annotator (rater 4), they still underperfomed when compared to the other experts on all the metrics. From these observations, they concluded that the detection and segmentation of new lesions in longitudinal studies is a challenging task even for experts. Therefore, a final mask reviewed by external experts may still be required for longitudinal analysis, even if U-net approaches can be considered the current state-of-the-art.}

\section{Conclusion and future directions}
\label{sec-conclusion}
\textcolor{black}{
We reviewed recent methodologies for MS lesion segmentation, with a special focus on deep-learning approaches. The analysis of the results from the public challenges held at MICCAI 2008, 2016 and 2021; and ISBI 2015 facilitated comparison and evaluation of some of the most relevant methods. Despite substantial progress, there is no specific segmentation method that matches the standard of human experts. For example, in the MICCAI 2016 database, the Dice score between expert raters was 0.782, while the best method~\cite{palladino2020unsupervised} obtained a Dice score of 0.591~\cite{mckinley2016nabla}. Additionally, due to the scarcity of clinical validation studies and the requirement for strict standardisation of image acquisition, existing automated MS lesion segmentation tools are not recommended routinely by field expert in clinical settings~\cite{wattjes20212021}. Furthermore, additional challenges, and possible future directions, for MS lesion analysis include:
}

\subsection{Multi-center Studies}
Private MRI databases are commonly used for designing state-of-the-art methods, some of which are not released to the research community.  Conversely, publically available datasets contain a relatively small number of cases that do not represent the spectrum of real-world clinical imaging variances. Curation and sharing a sufficient amount of brain MRI data to design and analyze automatic methods is hampered by both patient privacy concerns and local governance restrictions. Separately, variances in MRI scanners (field strength, vendor, model, software) and acquisition protocols generate a wide variation in the MRI characteristics of individual subjects in any dataset. In fact, our meta-analysis of the post-challenge submissions suggests a lower performance ceiling on the dataset obtained from two different centers. Techniques that adapt data from different sources and transfer that knowledge without additional training or labeling are therefore essential.  Large multi-center studies, which capture ‘real-world’ variability, provide an opportunity to validate such techniques.

Perhaps the greatest challenge for deep learning applications in medical imaging is a lack of expertly annotated data~\cite{zhou2021review,panayides2020ai}. Training a deep network for a visual task normally requires a large amount of training data to obtain the optimal outcome, and medical imaging annotation usually requires either a highly trained medical imaging analyst or an expert specialist physician/radiologist.  The majority of state-of-the-art work in the field is therefore focused on either unsupervised segmentation or semi-supervised fine-tuning with domain adaptation strategies. Two new and interesting directions that could ameliorate a lack of high-quality manual annotations are weak supervision, where coarse labels (for example, image-level labels such as the number of lesions rather than their segmentation) are provided; and the use of low quality or noisy labels (for example, labels acquired at a lower resolution)~\cite{papandreou2015weakly}. Together, these strategies potentially counter the lack of dense expert annotations to deliver an accurate and robust lesion segmentation.

\textcolor{black}{Finally, distributed learning has recently emerged as a solution to the problem of data sharing for large multi-centre studies. This novel research area focuses on the use of distributed training nodes (e.g., a model for each hospital) to maintain privacy within the center and to share model weights and knowledge in a safe and secure manner. There are two main approaches to distributed learning: using a central node that aggregates the models from the other local nodes (i.e., federated learning~\cite{kaissis2020secure}) or a decentralized system where weights are shared globally by all nodes (i.e., swarm learning~\cite{warnat2021swarm}). We believe that this is an interesting research direction to promote collaboration while following security and privacy guidelines and it will most likely explode in MS lesion segmentation in the coming years, as the tools mature.}

\subsection{Lesion Subtyping}
Future segmentation strategies with multi-modal MRI input data should focus on distinguishing between multiple sclerosis lesion subtypes (such as destructive black holes, actively demyelinating, remyelinating and slowly enlarging lesions); and their functional anatomic locations. Although total lesion volume and lesion activity (the number of new and enlarging lesions measured between timepoints) constitute the most frequently employed lesion biomarkers in clinical trials, automated lesion phenotyping will potentially improve understanding of MS and its progression in individual patients by categorizing the disease into different forms and patterns. In clinical practice, patients with a substantial total lesion load may have few or even no neurological symptoms, while others with only a small number/volume of lesions may have a severe disability. This so-called “clinico-radiological paradox” is reflected in weak correlations between clinical endpoints such as the Expanded Disability Status Scale (EDSS) score and total T2 lesion load in MS clinical trials~\cite{hackmack2012can}. Detailed MS lesion sub-typing could potentially delineate evidence of both damage or repair within lesions, overcome the clinic-radiological paradox, and provide specific structure-function impairment information that can be exploited for individualized disease progression monitoring~\cite{pontillo2021radiomics}.

Finally, as mentioned in Section~\ref{sec-bg-les}, contrast-enhancing lesions, which suggest active inflammation and breakdown of the blood-brain barrier, are currently detected using gadolinium-enhanced T1 images. However, some gadolinium compounds may accumulate in the brain with repeated exposure. The detection of active inflammatory lesions without the need for contrast agents is therefore a challenging but worthwhile task suited to deep-learning techniques with multi-modal MRI training datasets.

\bibliography{biblio_mc} 
\bibliographystyle{IEEEtran}


\end{document}